**ORIGINAL ARTICLE**

# Ensemble of Loss Functions to Improve Generalizability of Deep Metric Learning methods

Davood Zabihzadeh*[1], Zahraa Alitbi[2], Seyed Jalaleddin Mousavirad[1]

[1]Computer Engineering Department, Hakim Sabzevari University, Sabzevar, IRAN

[2] Computer Engineering Department, Engineering Faculty, Ferdowsi University of Mashhad (FUM), Mashhad, Iran

* Corresponding Author  d.zabihzadeh@hsu.ac.ir

## Abstract

Deep Metric Learning (DML) learns a non-linear semantic embedding from input data that brings similar pairs together while keeps dissimilar data away from each other. To this end, many different methods are proposed in the last decade with promising results in various applications. The success of a DML algorithm greatly depends on its loss function. However, no loss function is perfect, and it deals only with some aspects of an optimal similarity embedding. Besides, the generalizability of the DML on unseen categories during the test stage is an important matter that is not considered by existing loss functions. To address these challenges, we propose novel approaches to combine different losses built on top of a shared deep feature extractor. The proposed ensemble of losses enforces the deep model to extract features that are consistent with all losses. Since the selected losses are diverse and each emphasizes different aspects of an optimal semantic embedding, our effective combining methods yield a considerable improvement over any individual loss and generalize well on unseen categories. Here, there is no limitation in choosing loss functions, and our methods can work with any set of existing ones. Besides, they can optimize each loss function as well as its weight in an end-to-end paradigm with no need to adjust any hyper-parameter. We evaluate our methods on some popular datasets from the machine vision domain in conventional *Zero-Shot-Learning* (ZSL) settings. The results are very encouraging and show that our methods outperform all baseline losses by a large margin in all datasets. Finally, we develop a novel effective distance-based compression method that compresses both the coefficient and embeddings of losses into a single embedding vector. The embedding size of the proposed compressed method is identical to each baseline learner. Thus, it is fast as each baseline DML in the evaluation stage.

## 1. Introduction

The success of many machine learning tasks mainly depends on the distance or similarity measure utilized by them. The general distance metrics such as Euclidean often cannot capture the semantic relations between data items in input space. To this end, metric learning algorithms developed that aim to learn an optimal semantic distance function from data. Conventional metric learning methods focused on Mahalanobis distance learning that is equivalent to finding a *linear* transformation from the input to a semantic embedding space. Despite their success, they have two main limitations:

1- A linear transformation has a low capacity to learn semantic relations in many nonlinear datasets with complex class boundaries.
2- They cannot process complex data such as images and texts directly. Thus, we first should extract features from data and then forward the extracted features to the algorithm. Therefore, features extraction and metric learning process are performed independently and there is no guarantee to extract the best features for the task at hand.

In contrast, Deep metric learning (DML) methods perform both feature extraction and nonlinear semantic embedding learning simultaneously. Thus, they learn a better semantic embedding and surpass linear methods by a large margin. DML learns a nonlinear transformation function $f(x;\theta)$ from data that maps the input data $x$ from the input space (often image) to a similarity embedding space. Here, $\theta$ shows the parameters of both the deep feature extractor and embedding layer(s). In the learned embedding, the similar pairs should be close together whereas the dissimilar ones are kept far from each other.

A DML model consists of four main components: 1) deep feature extractor, 2) tuplet sampling, 3) embedding layer, and 4) loss function.

The $f(x;\theta)$ is often realized by a deep neural network where the typical Softmax layer at end of the network is replaced by a linear transformation layer named embedding. The embedding is a mapping matrix $W \in \mathbb{R}^{d \times e}$ that projects the extracted features of the deep model into e-dimensional embedding space. Many DML algorithms need training information in the form of doublets or triplets (tuples in general). To this end, many tuple sampling algorithms are developed to mine informative constraints from the input mini-batch.

Traditional DML methods (Chopra, Hadsell et al. 2005, Wang, Song et al. 2014, Hoffer and Ailon 2015) use Siamese or Triplet network shown in Figure 1. Instead, recent methods such as (Movshovitz-Attias, Toshev et al. 2017, Wang, Zhou et al. 2017, Qian, Shang et al. 2019, Jiang, Huang et al. 2020), get a mini-batch from the dataset and then forward it through the deep network. Subsequently, they sample pairs or triplets from the mini-batch before the loss layer.

The success of a DML algorithm greatly depends on its loss function. The *contrastive and triplet* are two seminal loss functions originally developed for the Siamese and Triplets networks, respectively. These loss functions deal with the fine-grained data to data relations among pairs or triplets. Recent works such as N-pair (Sohn 2016), Lifted Structure (Oh Song, Xiang et al. 2016), and Histogram loss (Ustinova and Lempitsky 2016) are introduced that better capture the global structure of embedding by considering more data relations simultaneously. Also, some work learns the embedding without a sampling procedure. For example, the clustering loss (Oh Song, Jegelka et al. 2017) directly optimizes the NMI[1]. Recently, *proxy-based* losses (Movshovitz-Attias, Toshev et al. 2017, Qian, Shang et al. 2019, Kim, Kim et al. 2020) attract much interest. They capture the structure of embedding by means of proxies that are learned from data jointly with other network parameters using BP[2]. These approaches do not need a tuple sampling step and have a higher convergence rate.

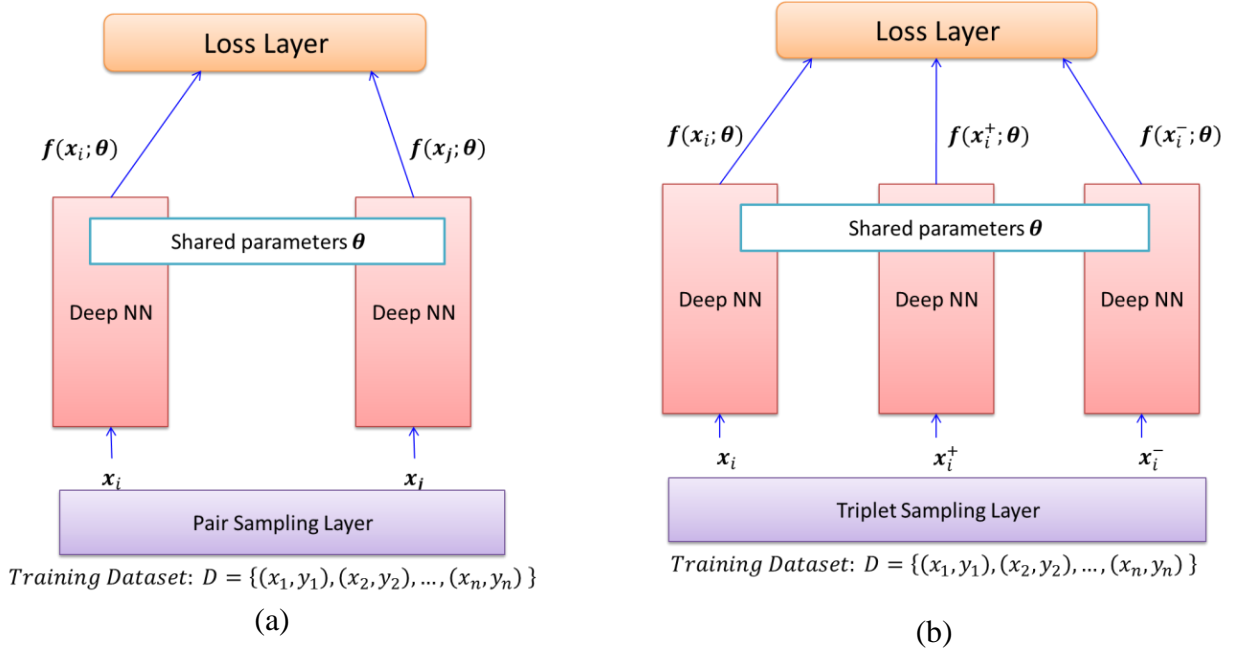

Figure 1- Architecture of (a) Siamese and (b) Triplet Networks

As seen, each loss function 1) has its merits, 2) deals only with some aspects of an optimal similarity embedding, and 3) there is no perfect loss. Besides, most applications of DML such as CBIR[3], person re-identification, and face recognition require that the learned metric generalize well on unseen categories during the evaluation phase. However, existing losses only concentrate to enhance the discrimination power of the learned embedding on the set of observed classes during the training stage.

To handle these issues, we propose novel methods that combine the losses on top of a shared deep network. Minimizing the ensemble of losses enforces the deep network to extract features that are consistent with all losses. Therefore, provided the selected losses be diverse and each focuses on different aspects of an optimal similarity embedding, the common semantic concepts that generalize well on new objects are learned by our methods.

---

[1] Normalized Mutual Information

[2] Back Propagation

[3] Content-Based Information Retrieval

The proposed methods can work with any set of existing losses. Since the loss's values may be at different ranges, we propose a novel rescaling approach based on EMA[1] to normalize them. Also, the proposed methods can optimize each loss function as well as its weight in an end-to-end paradigm without adjusting any additional hyper-parameter.

We develop some methods to combine the losses and evaluate them on some widely used datasets for image retrieval, clustering, and classification tasks. The results are very encouraging, and our methods outperform individual losses by a large margin in all experiments. Also, the best performance is obtained by the proposed *WEDL-DML*[2] method that considers a separate embedding layer per loss.

Finally, to reduce the retrieval time in the test phase, we develop a compression method named *WEDL-C* that compresses both the coefficients and embeddings of losses into a single embedding vector by means of a new *normalized distance-based* loss. The embedding dimension of *WEDL-C* is the same as baseline losses. Thus, it is fast as each baseline DML in the evaluation stage whereas its results considerably outperform all individual losses and are very competitive with that of *WEDL-DML*. In summary, the main contributions of this paper are summarized as follows:

1- We propose novel methods to build an ensemble of existing DML losses on top of shared deep network. Our methods enforce the deep network to extract features that are consistent with all losses and cover several aspects of an optimal semantic embedding simultaneously. Thus, the learned embedding generalizes well on unseen objects and classes in the evaluation stage.
2- The proposed methods can work with any set of existing losses Besides, we propose a new rescaling technique based on exponential moving average to normalize the losses values. Also, the proposed methods can optimize each loss function as well as its weight in an end-to-end paradigm without adjusting any additional hyper-parameter.
3- We investigate several approaches to implement the ensemble of losses and observe that the best performance is obtained when considering a separate embedding layer per loss.
4- A compression method named *WEDL-C*[3] is presented that compresses both the coefficients and embeddings of losses into a single embedding vector utilizing a new *normalized distance-based* loss.
5- Extensive experiments are performed on some challenging machine vision datasets for image retrieval, clustering, and classification tasks in a ZSL setting, and our methods surpass each baseline loss in all datasets by a large margin.

---

[1] Exponential Moving Average

[2] Weighted Ensemble of Diverse Losses-DML

[3] WEDL-Compact

The rest of the paper is organized as follows. Section 2 reviews existing DML work with a particular focus on loss functions. Section 3 presents the proposed ensemble of losses methods. Implementation details of each part of our methods are presented in Section 4. Section 5 reports the experiments and the obtained results. Finally, the conclusion of work along with recommendations for future work is given in Section 6.

## 2. Related Work

DML and its applications are discussed in detail in new surveys such as (Kaya and Bilge 2019). Here, we particularly focus on the loss functions used in different DML algorithms (Subsection 2.1). Our work is also relating to the ensemble of losses discussed in Subsection 2.2. Finally, Subsection 2.3 reviews existing work focused on the generalization of DML on unseen classes.

### 2.1 DML Loss Functions

Initially, (Chopra, Hadsell et al. 2005) used the Siamese network to learn a semantic embedding for the face verification task. The method aims to reduce the energy function between positive pairs and maximize it on negatives. The energy function is defined as Euclidean distance in the embedding space:

$$E_\theta = \|f(x_1; \theta) - f(x_2; \theta)\| \qquad (1)$$

For this purpose, the following *contrastive loss* is defined:

$$L(y, x_1, x_2, \theta,) = (1-y)\frac{2}{Q}(E_\theta)^2 + y2Q \exp\left(-\frac{2.77}{Q}E_\theta\right) \qquad (2)$$

where the binary variable y indicates whether the pair is positive ($y = 0$) or negative ($y = 1$). Also, $\theta$ indicates the network parameters. Loss functions such as (2) defined on a pair are called *contrastive*. The most widely used *contrastive* loss (Opitz, Waltner et al. 2018, Yuan, Deng et al. 2019, Li, Ng et al. 2020) is as follows:

$$L(y, x_1, x_2, \theta,) = \begin{cases} d, & \text{if } y = 0 \\ [\alpha - d]_+ = \max(0, \alpha - d), & \text{otherwise} \end{cases} \qquad (3)$$

where $d = d_\theta(x_1, x_2) \stackrel{\text{def}}{=} \|f(x_1; \theta) - f(x_2; \theta)\|_2^2$, and $\alpha$ is a margin.

The contrastive Binomial loss (Ustinova and Lempitsky 2016) is also adopted in some recent work (Ustinova and Lempitsky 2016, Opitz, Waltner et al. 2018, Chen and Deng 2019) and obtains encouraging results.

While contrastive losses are based on *absolute* distance values, *triplet* losses (Wang, Song et al. 2014, Hoffer and Ailon 2015, Ge 2018, Chen and Deng 2019, Yuan, Deng et al. 2019) concentrate on *relative* distances between positive and negative pairs. They also capture both similar and dissimilar pairs simultaneously. Thus, they often obtain a better performance in comparison with contrastive losses. The most popular triplet loss is margin-based Hinge loss defined as:

$$l((x_i, x_i^+, x_i^-)) = \begin{cases} 0, & if\ (d^- - d^+) \geq \alpha \\ \alpha - (d^- - d^+), & otherwise \end{cases} = [\alpha - (d^- - d^+)]_+ \quad (4)$$

where $d^+ = d_\theta(x_i, x_i^+)$ and $d^- = d_\theta(x_i, x_i^-)$ are Euclidean distances between positive and negative pairs in the embedding space, respectively.

Since the triplet loss deals with fine-grained data-to-data relation, it cannot capture the global structure of the semantic embedding well and has a low convergence rate. Many new losses are proposed to address these limitations.

The Angular loss (Wang, Zhou et al. 2017) constrains the angle at the negative point of the triangle formed by a triplet. It often achieves a better convergence rate compared to triplet loss and also is robust against feature variance.

Some work (Ni, Liu et al. 2017, Yao, She et al. 2020) extends the triplets to quadruplets. A quadruplet $(x_i, x_i^+, x_i^r, x_i^-)$ has an additional *relative* point and the loss it defined to enforce the following distance relation between points in the embedding space:

$$d_\theta(x_i, x_i^+) < d_\theta(x_i, x_i^r) < d_\theta(x_i, x_i^-)$$

Figure 2 illustrates the differences between contrastive, triplet, and quadruplet losses.

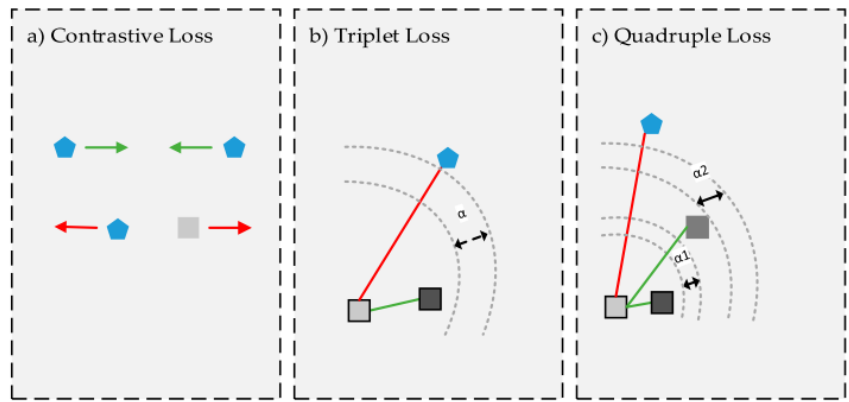

**Figure 2- Contrastive vs. Triplet vs. Quadruplet losses (Kaya and Bilge 2019)**

To better capture the global structure of semantic space, some losses optimize many data relations in a given mini-batch at a time. For example, n-pair loss (Sohn 2016) samples $N - 1$ negative points (one from each opposite class) for each positive pair. Then, it pushes all negative pairs away from the positive simultaneously.

$$l_{n-pair}((x, x^+, x_1^-, x_2^-, \ldots, x_{N-1}^-)) = -\ln\left(1 + \sum_{j=1}^{N-1} \exp(S_j^- - S^+)\right) \quad (5)$$

$where, S^+ = f(x; \theta)^T f(x^+; \theta), \quad S_j^- = f(x; \theta)^T f(x_j^-; \theta)$

Lifted Structure (Oh Song, Xiang et al. 2016) considers all possible negative constraints for each positive pair in the input mini-batch. The histogram loss (Ustinova and Lempitsky 2016) first estimates the similarity distributions of positive and negative pairs using histograms. Then, it minimizes the probability that a negative has a larger similarity score than a positive pair.

All discussed methods require a sampling algorithm to generate tuples (pair, triplet, etc.,) from a given dataset. The performance of the learned metric depends on the quality of mined tuples. To this end, many sampling algorithms such as *easy, hard, semi-hard*, and *n-pair* exist in the literature (Kaya and Bilge 2019). Sampling algorithms usually involve considerable time and storage cost. Hence, some DML methods (Rippel, Paluri et al. 2015, Oh Song, Jegelka et al. 2017) are presented to optimize the embedding without needing any sampling process. The magnet loss (Rippel, Paluri et al. 2015) maintains the distribution of classes in the embedding space. Then, it achieves local discrimination by penalizing overlaps between classes. The clustering loss (Oh Song, Jegelka et al. 2017) learns a semantic embedding by directly optimizing the clustering quality metric NMI[1].

Recently, proxy-based losses (Movshovitz-Attias, Toshev et al. 2017, Qian, Shang et al. 2019, Kim, Kim et al. 2020) are proposed that achieve a very promising performance as well as a high convergence rate. Proxy-NCA loss (Movshovitz-Attias, Toshev et al. 2017) replaces the positive and negative data points in a potential training triplet by a set of proxy centers learned jointly with other deep network parameters. Soft Triple (Qian, Shang et al. 2019) establishes a connection between triplet and normalized cross-entropy loss. It demonstrates that the normalized cross-entropy loss is identical to a smoothed version of triplet loss where each class is unimodal. Then, it develops the Soft Triple loss that extends the classification loss by modeling each class with multiple centers. Also, an effective regularization term is presented to prune surplus centers in each class. Proxy Anchor (Kim, Kim et al. 2020) considers proxies as anchor points and relates them to all data in an input mini-batch.

As seen, each loss function has its merits and considers different aspects of an optimal semantic embedding. This paper presents effective approaches to combine a set of losses to exploit the advantages of each and enhance the generalization of learned embedding on unseen classes.

### 2.2 Combining the losses

The ensemble of losses first is introduced in (Hajiabadi, Monsefi et al. 2019) for a regression task in a noisy environment. Here, the objective problem optimizes both losses and their coefficients using Half-Quadratic. The work is extended for neural networks and a classification task in (Hajiabadi, Babaiyan et al. 2020). Our methods extend this formulation for DML from various aspects. We resolve some issues by normalizing the loss values using EMA during optimization. Also, we enhance the performance of the ensemble by considering a separate embedding layer per loss and an additional diversity term.

### 2.3 Generalizability of DML

---

[1] Normalized Mutual Information

Most DML applications such as zero-shot (or few-shot) image classification, image retrieval, and person re-identification are challenging due to: 1) They have a large number of categories, 2) Some classes only contain a few examples or even zero in the training stage.

Most research in DML concentrates on increasing the discrimination power of learned embedding on a set of observed classes whereas neglecting the importance of generalization power of embedding on unseen classes in evaluation time.

MSML[1] (Jiang, Huang et al. 2020) alleviates the issue by extracting both coarse-level semantic and low-level visual features. Then, it computes the similarity of objects using learned multi-scale feature maps in an FSL[2] task.

(Li, Yu et al. 2020) aims to enhance the generalizability of DML by extending the triplet loss to K-tuplet loss. Here, the anchor point is associated with K negatives simultaneously. However, this approach is effective to capture the global structure of the learned embedding and observed in previous research such as N-pair (Sohn 2016).

(Chen and Deng 2019) introduces the *energy confusion* term that penalizes the distance between objects from two different classes. The confusion term is in contrast with one of the primary goals of DML: *keep examples of different categories far from each other*. However, the authors claim that the term can effectively reduce the overfitting problem on seen classes.

This paper also aims to enhance the generalization power of DML in a ZSL setting. To this end, it presents effective methods to build an ensemble of diverse loss functions on top of a shared feature extractor. That enforces the deep feature extractor to mine features that are consistent with all losses. Since the selected losses are diverse and each focuses on different aspects of an optimal semantic embedding, our effective combining methods result in considerable improvement over any individual loss and generalize well on unseen categories.

## 3. The Proposed Methods

### 3.1 Feature Extraction

Most DML applications such as image retrieval and person re-identification use image datasets. CNN models are very popular for these tasks. They can learn hierarchical patterns from input images. The initial modules of CNN extract general and small patterns from a given image and subsequent layers provide more semantic and discriminative concepts.

Many DML methods use the feature vector from the last hidden layer ($\boldsymbol{u} \in \mathbb{R}^l$) of a pre-trained CNN model. However, this representation increases the dependency on the observed classes and mainly focuses on specific regions from the input images. To address this issue, we adopt *General Discriminative Feature Learning (GDFL)* presented in our previous work (Al-Kaabi, Monsefi et al. 2021). GDFL divides the network into several modules and uses the output feature maps of selected intermediate modules to extract features. The feature maps of middle layers contain more general concepts; however, they are less discriminative. GDFL handles this problem by increasing the similarity between these features and the feature vector $u$. Figure 3 illustrates the process of GDFL.

---

[1] Multi-Scale Metric Learning

[2] Few-Shot Learning

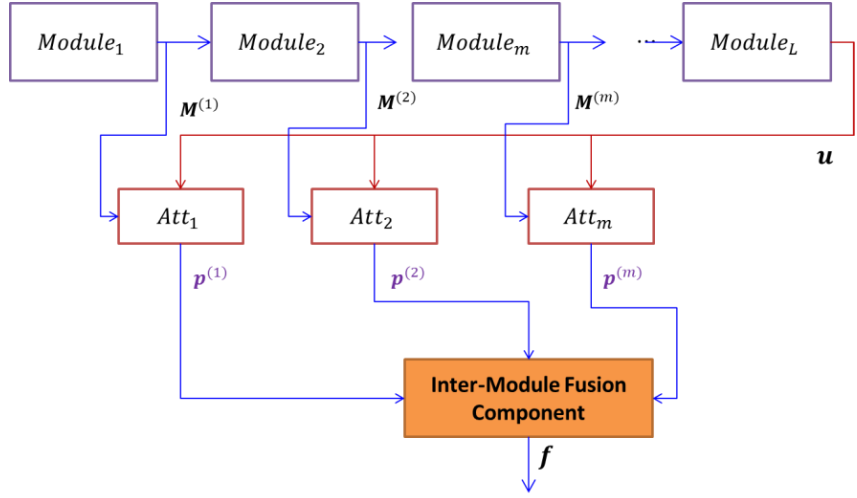

**Figure 3- The GDFL architecture (Al-Kaabi, Monsefi et al. 2021).**

Let $M^{(i)} \in \mathbb{R}^{C_i \times H_i \times V_i}$ be the feature maps from the i-th module. The $Att_i$ module attends $M^{(i)}$ to the discriminative feature vector $u$ and assigns an attention weight $a_j^{(i)}$ to each feature vector $q_j^{(i)} \in \mathbb{R}^{c_i}$, $j = 1,2,\ldots,(H_i \times V_i)$ in $M^{(i)}$. Subsequently, the attended channels are averaged to form general discriminative feature vector $p^{(i)}$ as follows:

$$p^{(i)} = \sum_{j=1}^{H_i \times V_i} a_j^{(i)} q_j, \qquad i = 1,2,\ldots,m$$

Here, we adopt the modified additive attention mechanism formulated as:

$$S(q_j, u) = W_i^T \sigma\left(W_i^{(1)} q_j + u\right) \qquad (6)$$

Extensive experimental results on some widely used machine vision datasets confirm that GDFL greatly enhances the performance of learned embedding in a ZSL setting.

### 3.2 Ensemble of Loss Functions

Let $\phi(.;\theta_f)$ denote a non-linear transformation function defined by the deep neural network model ($\theta_f$ indicates the network parameters). In training state, the model gets a mini-batch $B = \{((x_1, y_1), (x_2, y_2), \ldots, (x_N, y_N))\}$ from a training dataset:

$$\phi(B; \theta_f) = \{((\phi_1, y_1), (\phi_2, y_2), \ldots, (\phi_N, y_N))\}, \text{ where } \phi_i = \phi(x_i; \theta_f) \in \mathbb{R}^d.$$

The embedding layer is often implemented by a linear transformation $W \in \mathbb{R}^{d \times e}$ where d denotes the number of extracted features and $e$ represents the embedding dimension. Thus, the embedding representation of the input data $x_i$ is:

$$f(x_i, \boldsymbol{\theta} = \{\boldsymbol{\theta}_f, W\}) = \boldsymbol{\phi}_i^T W \tag{7}$$

Let $L = \{l_j\}_{j=1}^{M}$ represent the set of selected deep metric loss functions. Each loss $l_j$ receives $f(B; \boldsymbol{\theta})$ from the shared network and produces $l_j(f(B; \boldsymbol{\theta}))$. Note that some metric losses (such as the *contrastive*, *triplet*, and *N-pair*) require a sampling procedure to generate tuple constraints from the input batch. In this case, we initially sample tuples from $f(B; \boldsymbol{\theta})$ and then pass them to the loss function.

Since the losses are built upon a shared deep model, minimizing the ensemble of losses enforces the model to extract features that are consistent with all losses. Therefore, if the selected losses are diverse and emphasize different aspects of an optimal similarity embedding, the semantic concepts that generalize well on new objects are learned by the model. However, since the values of losses are at different scales, the naïve combination of them has a bias on the losses that produce large values. To overcome this issue, we propose to normalize them using a moving-average strategy. More precisely, let $\bar{L} = \{\bar{l}_j\}_{j=1}^{M}$ denote the current set of mean values of losses. At each training step, we compute the average of all means as:

$$\bar{l} = \frac{1}{M} \sum_{j=1}^{M} \bar{l}_j$$

Then, the normalized loss values are estimated as follows:

$$\widehat{l}_j = l_j \frac{\bar{l}}{\bar{l}_j} \tag{8}$$

After, we update the $\bar{L} = \{\bar{l}_j\}_{j=1}^{M}$ using the Exponential Moving Average (EMA).

$$\bar{l}_j^{(new)} = l_j \left(\frac{s}{1+k}\right) + \bar{l}_j \left(1 - \left(\frac{s}{1+k}\right)\right) \tag{9}$$

where $s$ is a smoothing factor, and $k$ indicates iteration number. The proposed objective function is formulated as follows:

$$l(f(B; \boldsymbol{\theta})) = \sum_{j=1}^{M} c_j \widehat{l}_j(f(B; \boldsymbol{\theta})), \quad \text{where} \sum_{j=1}^{M} c_j = 1 \text{ and } c_j \geq 0 \tag{10}$$

To combine the losses, A simple strategy is to set all coefficients $\{c_j\}_{j=1}^{M}$ identical (i.e., $c_j = \frac{1}{M}$). The results in the experiment section indicate that this simple strategy is indeed effective and resulting loss surpasses individual losses in the ZSL setting. A more sophisticated approach is to learn coefficients as well as network parameters through the BP[1] mechanism. To this end, we cast the provided objective problem into an equivalent unconstrained form as follows:

$$l(f(B;\boldsymbol{\theta})) = \sum_{j=1}^{M}(c_j^2 + \alpha)\hat{l}_j(f(B;\boldsymbol{\theta})) + \eta\left(\sum_{j=1}^{M}(c_j^2 + \alpha) - 1\right)^2 \quad (11)$$

Here, the parameter $\alpha$ maintains a minimum weight for each loss function. We set $\alpha = \frac{1}{4M}$ throughout the experiments. The $\eta$ must be a significantly large value to ensure the constraint $\sum_{j=1}^{M}(c_j^2 + \alpha) - 1$ is held at each network update. We consider $\eta = 100$ in the experiments. Figure 4 illustrates the overall training steps of the proposed model named WEL-DML[2].

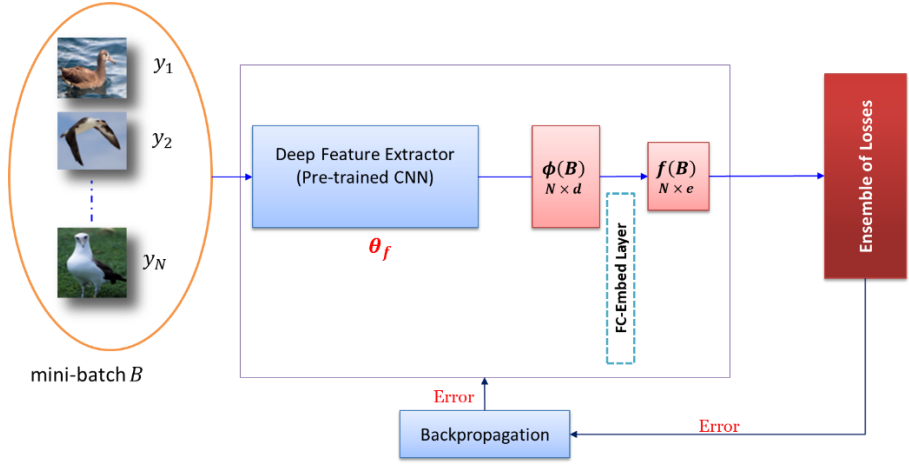

**Figure 4- The overall training process of the proposed ensemble of losses**

### 3.3 Diversity Enhancement

To further enhance the diversity among losses, we propose to consider a separate embedding layer per loss. It also promotes the representation capacity of the embedding. In this way, the loss produced by each learner only updates the corresponding embedding layer plus the shared deep feature extractor. We named this method WEDL-DML[3]. The training process of WEDL-DML is depicted in Figure 5. As see, the $j^{th}$ embedding layer is implemented by a linear transformation matrix $W_j \in \mathbb{R}^{d \times e}$ that maps the extracted feature vector into the embedding space.

---

[1] Back Propagation

[2] Weighted Ensemble Losses DML

[3] Weighted Ensemble of Diverse Losses-DML

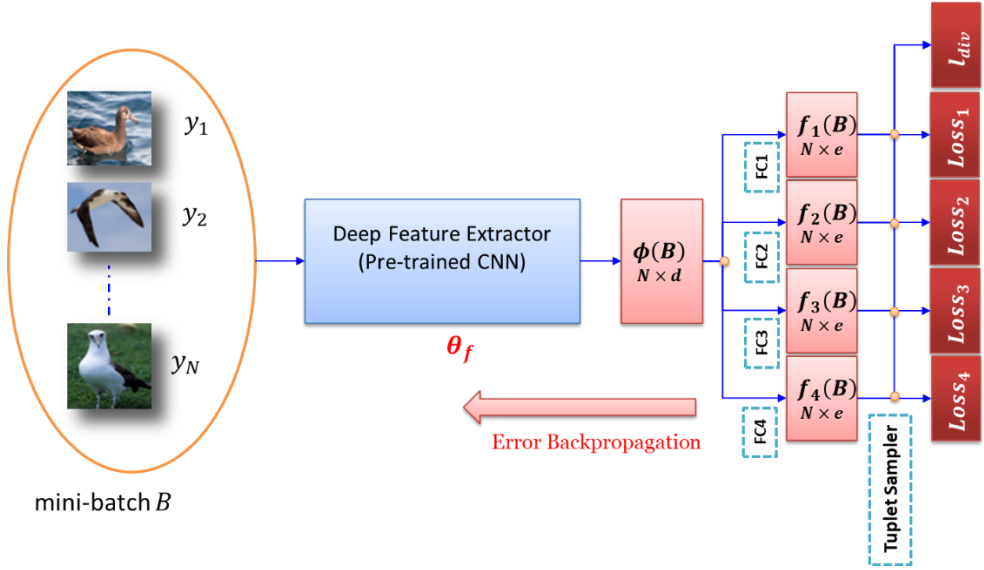

Figure 5- The training process of the proposed WEDL-DML(m=4)

To encourage diversity among embeddings, we also propose a new distance-based diversity term as follows. First, we normalize the embedding vectors into *unit norm length* (i.e., $\|f_j(x_i; \theta_j = \{\theta_f, W_j\})\| = 1$) and compute the mean of pairwise distances between any pair of learners in the given mini-batch.

$$D = \frac{2}{M(M-1)} \sum_{j=1}^{M} \sum_{k=j+1}^{M} \left( \sum_i \|f_j(x_i) - f_k(x_i)\| \right) \qquad (12)$$
$$= \frac{2}{M(M-1)} \sum_{j=1}^{M} \sum_{k=j+1}^{M} \left( \sum_i 2 - f_j(x_i)^T f_k(x_i) \right)$$

Note that the maximum distance between each normalized pair $(f_j(x_i), f_k(x_i))$ is 4. We choose 2 as a threshold (margin) and propose the following diversity term:

$$l_{div} = \max\{0, 2 - D\} \qquad (13)$$

Finally, we can formulize the optimization problem of this approach as:

$$l\left(\{f_j(B;\theta_j)\}_{j=1}^{M}\right) = \sum_{j=1}^{M}(c_j^2 + \alpha)\hat{l}_j\left(f_j(B;\theta_j)\right) + \eta \left(\sum_{j=1}^{M}(c_j^2 + \alpha) - 1\right)^2 + \lambda l_{div} \qquad (14)$$

After training the model with the above ensemble of loss functions, in the evaluation stage, we compute the distance between two data points $x$ and x' at evaluation time as follows:

$$d(x, x')^2 = \sum_{j=1}^{M}(c_j^2 + \alpha)\|f_j(x) - f_j(x')\|_2^2 = \sum_{j=1}^{M}(c_j^2 + \alpha)\left(2 - f_j(x)^T f_k(x')\right) \qquad (15)$$

where the $(c_j^2 + \alpha)$ is the coefficient of $\hat{l}_j$ learned at the training stage. The proposed WEDL-DML method provides a substantial performance improvement over any baseline DML (learned by each loss individually). However, computing the distance using equation (15) demands more time and memory to some extent.

### 3.4 Compress the Model

Here, we proposed a compact form of WEDL-DML named *WEDL-C*[1] that compresses the coefficient and embeddings $\{(c_j^2 + \alpha), f_j(.)\}_{j=1}^{M}$ into a single embedding vector $g(.)$. To this end, we design a simple regressor that intends to learn the distance function (15). The architecture of the regressor is shown in Figure 6. First, the weighted embeddings are concatenated with each other and form $f_{con}$ embedding. The concatenation weights are equals to $\left\{\sqrt{(c_j^2 + \alpha)}\right\}_{j=1}^{M}$. Note that the $d(x, x')^2$ in (15) is equal to $\|f_{con}(x) - f_{con}(x')\|^2$. Then a non-linear transformation with *tanh* activation function is applied to $f_{con}$ and compresses it into the $g(x) \in \mathbb{R}^e$.

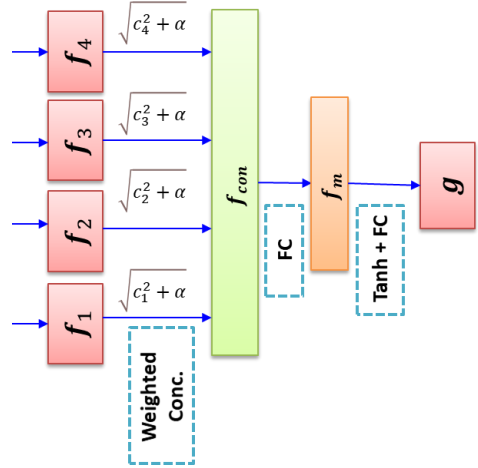

**Figure 6- Architecture of the regressor utilized to compress the embeddings (m=4)**

The regressor is trained by the proposed *normalized distance loss* function as follows. First, we compute the distance between any pair $(x_i, x_j)$ in the input mini-batch B and form the following normalized distance matrix:

$$\mathcal{K} = \frac{1}{\sum_{i,j=1}^{N} d_{ij}} \mathcal{D}, \quad \text{where } d_{ij} = d(x_i, x_j)^2 \text{ according to (15)}$$

Similarly, the normalized distance matrix of the *output* of the regressor is calculated:

$$\mathcal{K}' = \frac{1}{\sum_{i,j=1}^{N} d_{ij}} \mathcal{D}', \quad \text{where } d'_{ij} = \left\|g(f_{con}(x_i)) - g(f_{con}(x_j))\right\|_2^2$$

Finally, the normalized distance loss is defined as:

---

[1] WEDL-Compact

$$l_{dist} = \frac{1}{N^2} \|\mathcal{K} - \mathcal{K}'\|_{fro}^2 \tag{16}$$

Hence, by minimizing the distance loss, the compressed embedding learns the distance function obtained by the more powerful WEDL-DML method.

## 4. Implementation Details

We implement our proposed methods by following previous work such as (Movshovitz-Attias, Toshev et al. 2017, Qian, Shang et al. 2019) and utilize the pretrained BN-Inception. The size of the feature vector of the last hidden layer is 1024 (i.e., $\boldsymbol{u} \in \mathbb{R}^{1024}$). To extract features, we apply *GDFL* to BN-Inception and utilize the output feature maps of modules *inception_4d_output, inception_4e_output, and inception_5a_output*. We attend these feature maps to $\boldsymbol{u}$ using the modified additive attention mechanism (6). Figure (8) illustrates the simplified architecture of the deep feature extractor using GDFL.

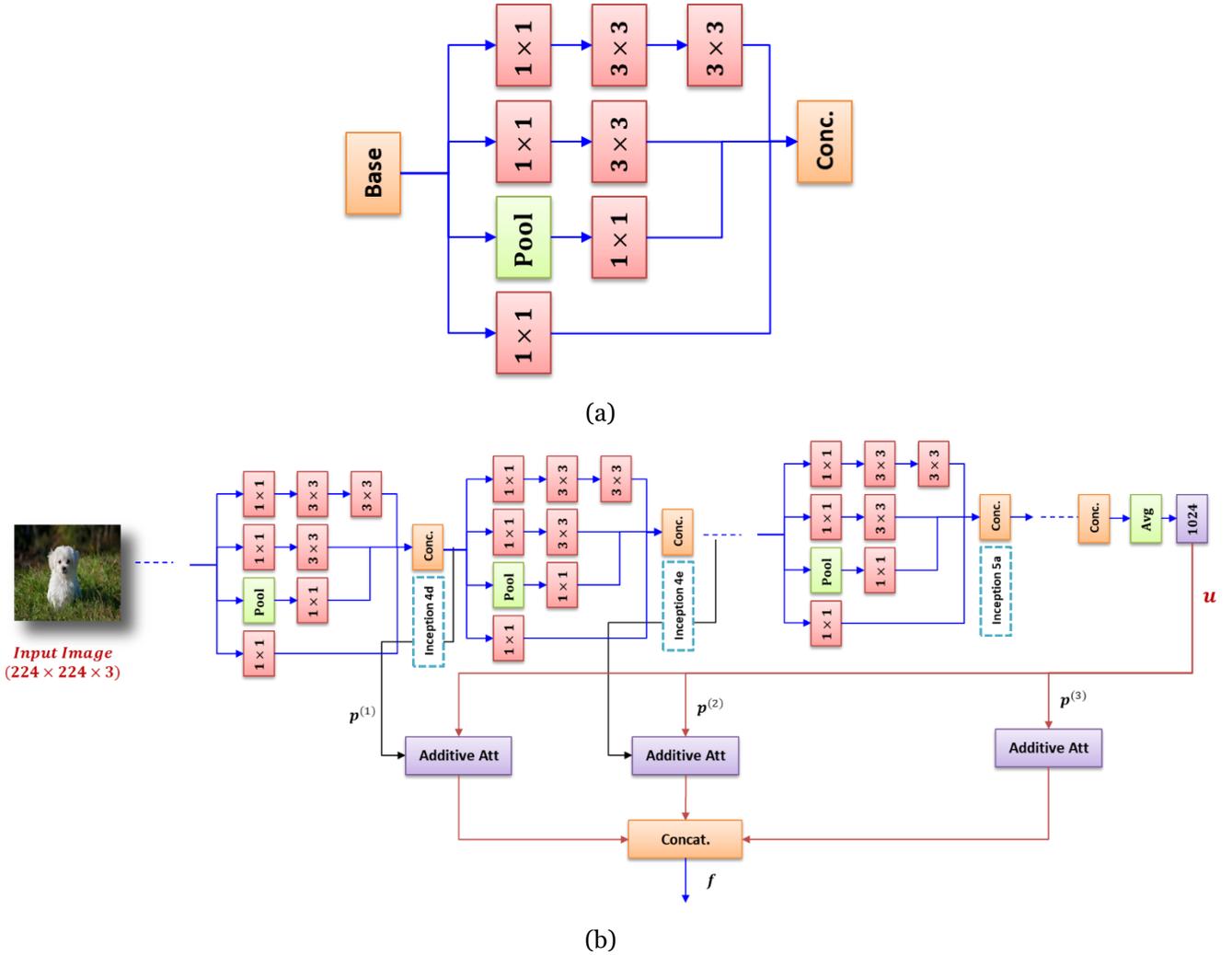

**Figure 7- (a)-The architecture of each Inception module in the *BN-Inception*, (b) The Simplified architecture of the deep feature extractor using GDFL (Al-Kaabi, Monsefi et al. 2021).**

We use Adam as the optimization method (with learning rate: $10^{-4}$, $\epsilon$: 0.01, and weight decay $10^{-4}$) in all experiments. Also, the embedding layer(s) learning rate is set to $10^{-3}$ (10 times faster than other layers). Our work is implemented by Pytorch and the source code is publicly available at https://github.com/d-zabihzadeh/DML-Ensemble-of-Losses.

To preprocess the input images, we follow previous work (Movshovitz-Attias, Toshev et al. 2017) and resize the images to $256 \times 256$. Then, we subtract the mean of the ImageNet dataset from the image. In the training stage, we crop a random $224 \times 224$ patch from the image and forward the path through the network. In the test phase, we use the $224 \times 224$ center crop of an image to extract the final feature embedding.

### 4.1 Selected Loss Functions

Loss functions are the central parts of any DML algorithm. Each loss function has its own merits and emphasizes some aspects of optimal semantic embedding. For example, *contrastive* and *triplet-hinge* losses deal with fine-grained data-to-data relations. On the other hand, recent proxy-based losses such as (Movshovitz-Attias, Toshev et al. 2017, Qian, Shang et al. 2019) capture the global structure of the embedding. Our ensemble of losses approach can exploit the advantages of each. Here, we select four different diverse loss functions as follows:

I. **Triplet-Hinge Loss**: This function formulated in (3), is a margin enhancing loss that leads to a better generalization on unseen data. It also deals with relative distances instead of absolute ones.

II. **Binomial Loss (Ustinova and Lempitsky 2016)** is a *normalized* variant of *contrastive* loss that has a *smooth* gradient shown in Figure 8. Let *s* be the cosine similarity in embedding space defined as:

$$s(f(x), f(x)') = \frac{< f(x), f(x)' >}{\|f(x)\|\|f(x')\|}$$

The value of *s* is between $[-1, 1]$. The *binomial deviance loss* is defined as:

$$l_{BD}(s, y) = \begin{cases} \log(1 + e^{-\beta_1(s-\beta_2)}), & y = 1 \text{ (similar pair)} \\ \log(1 + e^{\beta_1(s-\beta_2)C}), & \text{otherwise} \end{cases} \quad (17)$$

where $\beta_1$ and $\beta_2$ are scaling and translation hyper-parameters and set to 2 and 0.5, respectively (similar as (Ustinova and Lempitsky 2016)). The hyper-parameter C balances the loss for positive and negative pairs and is adjusted equal to 25 for negative pairs as (Ustinova and Lempitsky 2016).

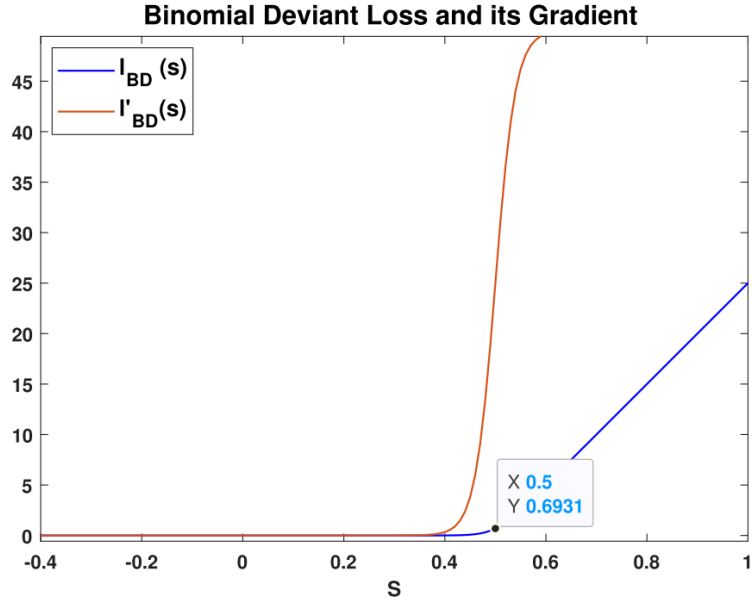

Figure 8- Binomial deviance loss and its smoothed gradient for negative pairs ($y \neq 1$)

III. **Proxy NCA (Movshovitz-Attias, Toshev et al. 2017):** This loss function can learn the semantic embedding without sampling strategy by means of proxies learned from data in the training stage.

Let $P$ be the set of proxies. For any data element $x$, the proxy of $x$ (denoted by $p(x)$) is defined as:

$$p(x) = \arg\min_{p \in P} d(x, p)$$

The Proxy NCA loss over the potential triplets $(x, y, Z)$ and a distance function $d(.,.)$ is formulated as:

$$l_{Proxy-NCA}(x, y, Z) = -\log \frac{\exp\left(-d(x, p(y))\right)}{\sum_{p(z) \in Z} \exp\left(-d(x, p(z))\right)} \tag{18}$$

This algorithm has a prominent convergence rate and captures the global structure of the embedding via proxies well. The experimental results show that Proxy NCA consistently outperforms other selected losses on evaluated datasets.

IV. **Classification Loss**: Here, we aim to learn the embedding using a simple classifier trained on top of the shared feature extractor. We choose the label smoothing approach to train a robust classifier. Label smoothing replaces a target label $y_{hot}$ (in one-hot encoding format) with a noisy label $y^s$ by combining $y^h$ with a uniform distribution as follows:

$$y^s = (1 - \gamma)y^h + \frac{\gamma}{C} \tag{19}$$

where $\gamma$ is a smoothing factor and $C$ is the number of available classes. Label smoothing addresses both overfitting and overconfidence problems well. The architecture of the classifier is shown in Figure 9.

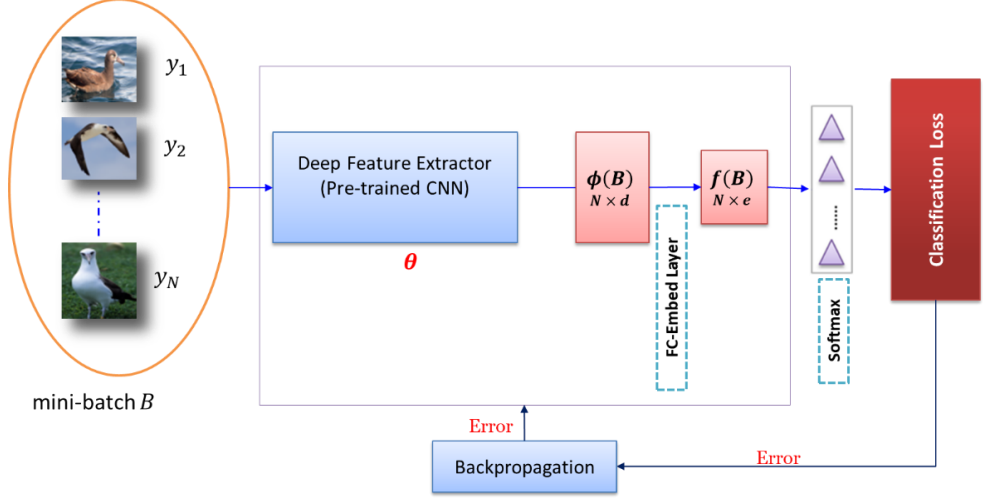

**Figure 9- Architecture of the Classifier**

The input of the classifier is the extracted feature vector $\boldsymbol{\phi}$. The classifier consists of a simple embedding layer followed by a Softmax layer. The smoothed classification loss is:

$$l_c(\boldsymbol{x}, \boldsymbol{y}_s) = -\sum_{k=1}^{C} y_k^s \log[\boldsymbol{h}(\boldsymbol{f}; \boldsymbol{\theta}_c)]_k \tag{20}$$

where $\boldsymbol{f}$ is the embedding of the $\boldsymbol{x}$ and $[\boldsymbol{h}(\boldsymbol{f}; \boldsymbol{\theta}_c)]_k$ shows output $k$ of the classifier.

## 5. Experimental Results

We evaluate our work in a ZSL setting on three popular image retrieval datasets. Here, we study how much the proposed ensemble of losses methods improve the performance of the learned embedding for image retrieval and classification tasks. Subsequently, ablation study and hyper-parameter analysis of the proposed methods are provided.

### 5.1 Image Datasets

The Oxford 102 Flowers, CUB-200-2011, and CARS-196 datasets are three popular datasets used in our experiments. The we evaluate our work on these datasets in a ZSL setting where the train and test classes are disjoint. More precisely, we divide the available $C$ classes in a dataset by two. The first set (range $(0, C/2)$) is utilized for training, and the remaining classes are left for evaluation. Table 1 summarizes the specifications of these datasets.

Table 1- The Specifications of selected image datasets in our experiments.

| Data Set | #classes | #samples | Train Set | Test Set |
|---|---|---|---|---|
| Oxford Flowers-102 (Nilsback and Zisserman 2008) | 102 | 6,552 | The first 51 classes in range(0,51) (including 2,807 images) | The last 51 classes (in range(52,102) containing 3,745 images) |
| CUB-200-2011 (Wah, Branson et al. 2011) | 200 | 11,788 | The initial 100 classes in range (0,100) (including 5,864 images) | The remaining 100 classes (in range (101,200) containing 5,924 images) |
| CARS-196 (Krause, Stark et al. 2013) | 196 | 16,185 | The first 98 classes in range (0,98) (including 8,054 images) | The last 98 classes (in range (99,196), containing 8,131 images) |

## 5.2 Evaluation Metrics

The standard metrics in information retrieval and clustering are adopted to evaluate the competing methods. The metrics include Recall@$k$ ($k = 1,2,4,8$), and NMI[1].

Recall@$k$ indicates the proportion of relevant images in the top-$k$ retrieved objects whereas NMI measures the quality of clustering. Let $C = \{c_1, c_2, \ldots, c_n\}$ denote the clustering pseudo labels obtained by a clustering algorithm such as k-means. Given the targets $Y = \{y_1, y_2, \ldots, y_n\}$, NMI is calculated as:

$$NMI = 2 \times \frac{I(Y;C)}{H(Y) + H(C)} \tag{21}$$

where $I(Y;C)$ denotes the mutual information between $Y$ and $C$, and $H(.)$ is the entropy.

Also, we compute the accuracy of the kNN ($k=3$) in the ZSL setting. Here, a test image is classified correctly provided *2 or 3* relevant (with the same label) images be in the *3-top* retrieved images.

## 5.3 Experimental Setup

For all evaluated methods, we utilize the pretrained BN-Inception network. The GDFL is used to extract features from input images in the proposed methods. The embedding dimension is set to 64 like (Movshovitz-Attias, Toshev et al. 2017).

To adjust the hyperparameters of the models, we randomly select 15% of the training set as validation and perform a grid search. Since we aim to evaluate to performance of DML methods on unseen classes, we did not perform an exhaustive validation search to adjust hyper-parameters of baseline DML methods and almost retain their default values.

More precisely, we train all models using the Adam optimizer with an initial learning rate selected from $\{10^{-4}, 10^{-3}, 10^{-2}\}$. We decreased the learning using the cosine annealing scheduler. For Triplet loss, the margin is chosen from {0.01, .1, 1} and, the semi-hard sampling strategy is used for all datasets. For Binomial Loss, we retain the default values of hyper-parameters (i.e., $\beta_1 = 2, \beta_2 = 0.5, and\ C = 25$). For Proxy NCA, the learning rate of the DML layer is chosen from the set {0.01, 0.15}. Also, the smoothing factor of classification loss is set to 0.15. Finally, the coefficient of diversity term ($\lambda$) in the proposed methods is set to 0.01 in all experiments. Table 2 reports the hyper-parameters of DML algorithms as well as their adjustments.

---

[1] Normalized Mutual Information

Table 2- Hyperparameters of DML methods and their adjustments

| DML Method | Hyper-parameters |
|---|---|
|  | $model\text{-}lr: \{10^{-4}, 10^{-3}, 10^{-2}\}$ |
|  | $embed\text{-}size: 64,\ embed\text{-}lr: 10 \times model\text{-}lr$ |
| **Triplet Loss:** | $margin: \{.01, .1, 1\}$ |
|  | $sampling\ strategy:\ semi\text{-}hard$ |
| **Binomial Loss:** | $\beta_1: 2, \beta_2: 0.5,$ |
|  | $C: 25$ |
| **Proxy NCA:** | $dml\text{-}lr:\ \text{Adjusted from}\ \{.01, .15\}$ |
| **Classification Loss:** | $\gamma: 0.15, lr: .01$ |
| **WEDL-DML:** | $\lambda: 0.01$ |

## 5.4 Results and Analysis

We evaluate the performance of the proposed ensemble of losses methods as well as each selected DML loss individually. The evaluated datasets and their evaluation protocols are described in Table 1. The results are reported in Table 3, Table 4, and Table 5. Also, the test NMI and Recall@1 of the competing methods vs. epochs are depicted in Figure 10.

Table 3- Information Retrieval Results in a ZSL Setting on the Oxford 102 Flowers dataset.

| Method | Recall@1 | Recall@2 | Recall@4 | Recall@8 | NMI | kNN-Acc |
|---|---|---|---|---|---|---|
| **Triplet Loss** | 81.55 | 88.89 | 93.59 | 96.58 | 70.16 | 82.22 |
| **Binomial Loss** | 85.21 | 91.24 | 95.03 | 97.33 | 72.05 | 86.09 |
| **Proxy-NCA** | 86.3 | 91.24 | 95.33 | 97.38 | 73.79 | 88.38 |
| **Classification Loss** | 85.74 | 91.13 | 94.63 | 97.01 | 68.10 | 86.09 |
| **WEL-DML** | 91.43 | 94.87 | 97.09 | 98.34 | 77.10 | 91.67 |
| **WEDL-DML** | **94.23** | **96.37** | **97.84** | **98.88** | **82.35** | **94.13** |
| **WEDL-C** | 92.76 | 96.05 | 97.89 | 98.72 | 80.26 | 93.11 |

The obtained results are very encouraging, and the proposed methods consistently outperform all baseline DML methods by a large margin. Also, based on the results, we can conclude that:

1- Proxy-based DML methods such as Proxy-NCA capture better the global structure of the semantic embedding and also have a higher convergence rate.
2- Combining the several losses via the proposed approach is indeed beneficial and improves the performance of DML in a ZSL setting on challenging datasets. Also, it shows that the selected losses are diverse and deals with different aspects of an optimal semantic embedding.
3- WEDL-DML considerably performs better than WEL-DML in all evaluated datasets. Thus, considering a separate embedding layer per loss yields a better distance function.
4- The results obtained by WEDL-C are encouraging and the WEDL-C learns a better representation as compared to WEL-DML. It suggests that we can learn a better single semantic embedding from multiple losses by initially learn a separate embedding layer and coefficient weight per loss and then compress the results.
5- The proposed $l_{dist}$ (*normalized distance loss*) is indeed effective and by minimizing the loss, we can compress multiple embeddings and weights into a single powerful embedding efficiently.

Table 4- Information Retrieval Results in a ZSL Setting on the CUB-200-2011 dataset.

| Method | Recall@1 | Recall@2 | Recall@4 | Recall@8 | NMI | kNN-Acc |
|---|---|---|---|---|---|---|
| **Triplet Loss:** | 45.39 | 58.86 | 70.41 | 80.44 | 58.17 | 49.14 |
| **Binomial Loss:** | 55.03 | 66.95 | 77.67 | 86.26 | 63.40 | 57.09 |
| **Proxy-NCA:** | 54.02 | 66.83 | 78.07 | 85.7 | 63.62 | 56.74 |
| **Classification Loss** | 51.27 | 63.79 | 75.03 | 83.32 | 61.29 | 53.81 |
| **WEL-DML** | 57.38 | 69.21 | 79.1 | 86.72 | 64.69 | 60.04 |
| **WEDL-DML** | **61.09** | **72.81** | **81.35** | **88.17** | **66.84** | **63.83** |
| **WEDL-C** | 59.25 | 70.71 | 80.11 | 87.27 | 66.88 | 61.87 |

Table 5- Information Retrieval Results in a ZSL Setting on the CARS-196 dataset.

| Method | Recall@1 | Recall@2 | Recall@4 | Recall@8 | NMI | kNN-Acc |
|---|---|---|---|---|---|---|
| **Triplet Loss:** | 44.13 | 57.51 | 70.1 | 79.79 | 52.50 | 46.28 |
| **Binomial Loss:** | 58.85 | 69.84 | 79.9 | 87.32 | 56.31 | 60.85 |
| **Proxy-NCA:** | 65.67 | 76.67 | 84.68 | 90.65 | 61.20 | 67.78 |
| **Classification Loss** | 62.70 | 73.95 | 82.47 | 89.02 | 59.09 | 64.90 |
| **WEL-DML** | 69.88 | 79.47 | 86.48 | 91.71 | 62.45 | 71.71 |
| **WEDL-DML** | **76.04** | **84.25** | **89.79** | **93.65** | **66.53** | **77.32** |
| **WEDL-C** | 73.95 | 82.61 | 89.47 | 93.72 | 65.44 | 75.54 |

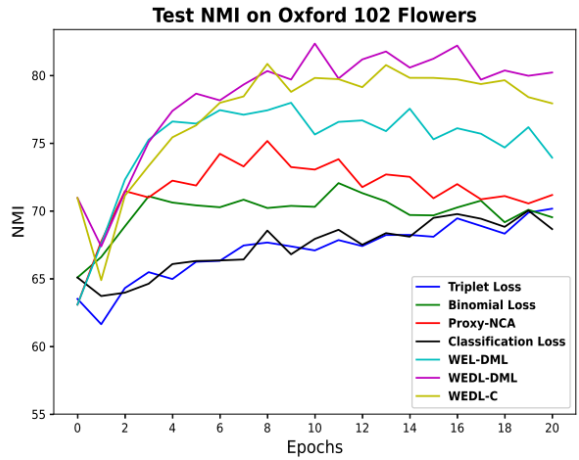
(a)

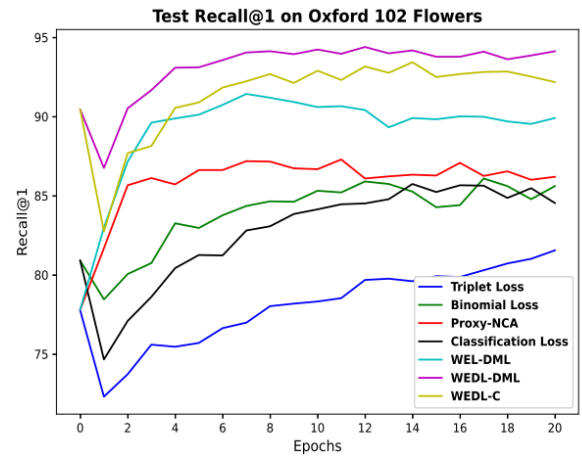
(b)

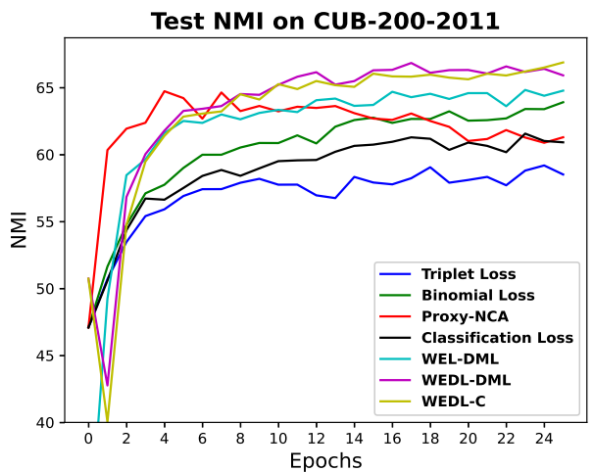
(c)

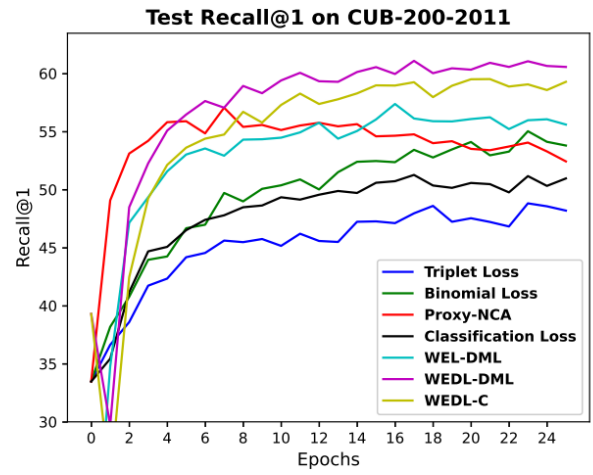
(d)

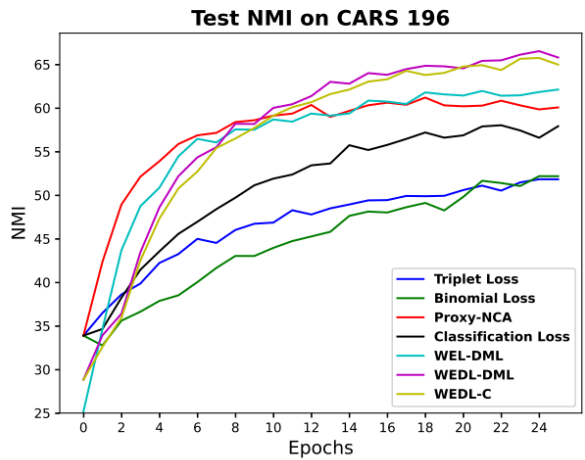
(e)

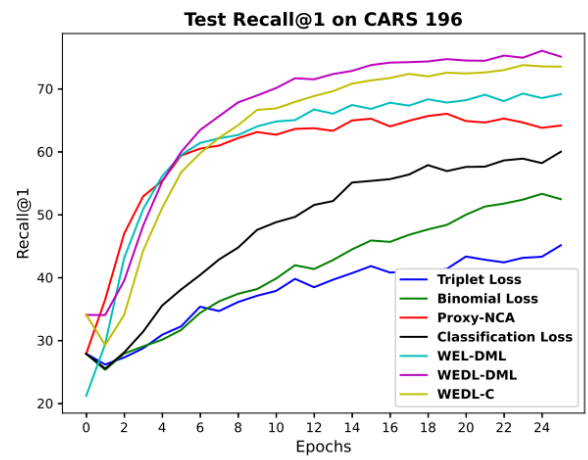
(f)

**Figure 10- NMI and Recall@1 of the evaluated methods on the Oxford 102 Flowers, CUB-200-2011, and CARS-196 datasets.**

## Ablation Study

In this experiment, we investigate the effectiveness of both *GDFL* and *Coefficient weights* by means of an ablation study. More precisely, we obtain two reduced variants for both the proposed methods WEDL-DML and WEDL-C.

1. $WEDL - DML^{(-GDFL)}, WEDL - C^{(-GDFL)}$, [1]: indicate WEDL-DML and WEDL-C *without* using GDFL, respectively. Here, we only use the output of the last hidden layer in the deep network as features.
2. $WEDL - DML^{(-CL)}, WEDL - C^{(-CL)}$: WEDL-DML and WEDL-C *without* learning the coefficient weight for the losses. Here, we consider equal weights (i.e., $c_j = \frac{1}{M}, j = 1, 2, \dots M$) for all losses.

We choose the Oxford 102 Flowers dataset in the ZSL setting (see Table 1) and compare the performance of the reduced methods with the original ones and the best loss (Proxy-NCA) by training the model for 20 epochs. The results are reported in Table 6. Also, Figure 11 shows the test NMI and Recall@1 of the evaluation methods as a function of the epoch.

Table 6- Information Retrieval Results of Ablation Study on the Oxford 102 Flowers dataset.

| Method | Recall@1 | Recall@2 | Recall@4 | Recall@8 | NMI | kNN-Acc |
|---|---|---|---|---|---|---|
| Proxy-NCA (Best Loss) | 86.3 | 91.24 | 95.33 | 97.38 | 73.79 | 88.38 |
| $WEDL - DML^{(-GDFL)}$ | 91.72 | 95.35 | 97.12 | 98.53 | 78.76 | 92.12 |
| $WEDL - DML^{(-CL)}$ | 94.05 | 96.56 | 98.05 | 98.88 | **81.69** | 93.99 |
| **WEDL-DML** | **94.47** | **97.06** | **98.18** | **99.04** | 81.42 | **94.39** |
| $WEDL - C^{(-GDFL)}$ | 89.69 | 93.99 | 96.88 | 98.4 | 77.02 | 90.68 |
| $WEDL - C^{(-CL)}$ | 93.24 | 96.29 | 97.73 | 98.85 | 78.26 | 92.98 |
| **WEDL-C** | **93.54** | **96.37** | **97.92** | **98.93** | **80.97** | **93.43** |

As the result indicate, GDFL is indeed effective in the ZSL setting and enhances the performance of both proposed methods. However, without GDFL, our methods still surpass the best loss (Proxy-NCA) consistently. It confirms that the proposed combination of losses improves the generalization of the learned embedding irrespective of feature learning.

Besides, the coefficient weights learned by our methods improve the results to some extent. Also, without coefficient learning, still, the combination of losses is effective and outperforms individual losses. The learned coefficient weights of WEDL-DML and WEDL-C in this experiment are as follows:

**$c_1$ (Triplet Loss): 0.10, $c_2$ (Binomial Loss): 0.30, $c_3$ (Proxy-NCA): 0.39, $c_4$ (Classification Loss): 0.21.**

---

[1] General Discriminative Feature Learning

It shows that the learning algorithm assigns the proper weights to losses and the coefficients correspond to the performance of DML losses (i.e., the more powerful loss is assigned a higher weight).

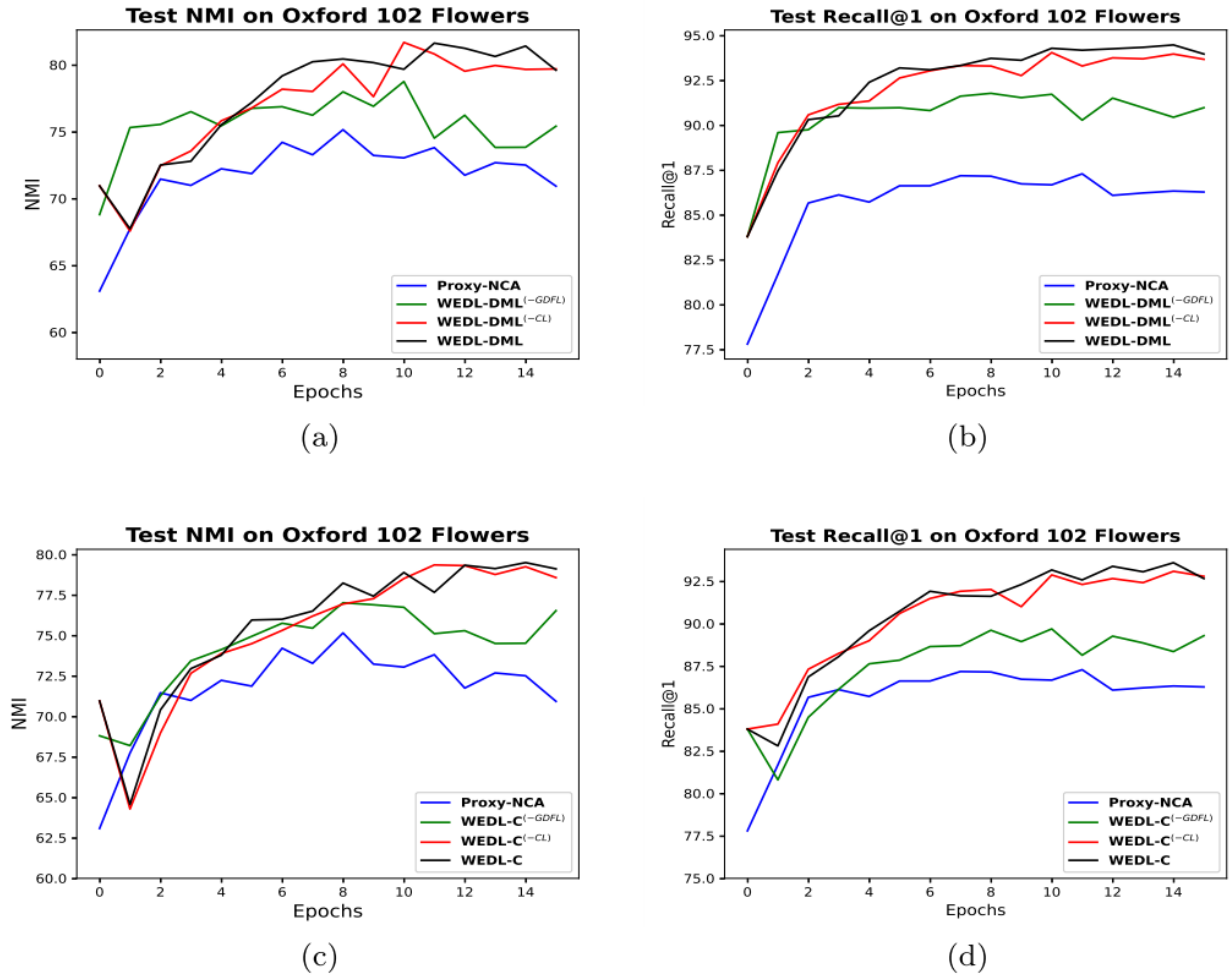

Figure 11- Test NMI and Recall@1 of the evaluated methods in Ablation Study on the Oxford 102 Flowers datasets

**Hyper Parameters Analysis**

The only hyper-parameter of our ensemble approach is the coefficient of diversity term ($\lambda$). In this experiment, we analyze the effect of the diversity term by training WEDL-DML at different $\lambda$ values. In this experiment, we choose the Oxford 102 Flowers dataset in the ZSL setting (see Table 1). Figure 12 depicts the NMI and Recall@1 of the WEDL-DML vs. $\lambda$ values on unseen categories.

As the results indicate, the proposed WEDL-DML has less sensitivity to the diversity term and works well at different $\lambda$ values. As the value of $\lambda$ increases, we observe a minor improvement of both NMI and Recall@1 metrics. Based on the results, we can conclude that:

1- For all evaluated $\lambda$ values, the proposed WEDL-DML outperforms the best loss function (Proxy-NCA) by a large margin.
2- The selected loss functions produce diverse embedding vectors. Hence, the diversity term only causes minor enhancement on the performance of WEDL-DML.

3- The Hyperparameter $\lambda$ is the only adjustable parameter in the proposed methods. As seen, the results are less dependent on the $\lambda$ values. Thus, we expect that our methods generalize well on new datasets and similar applications

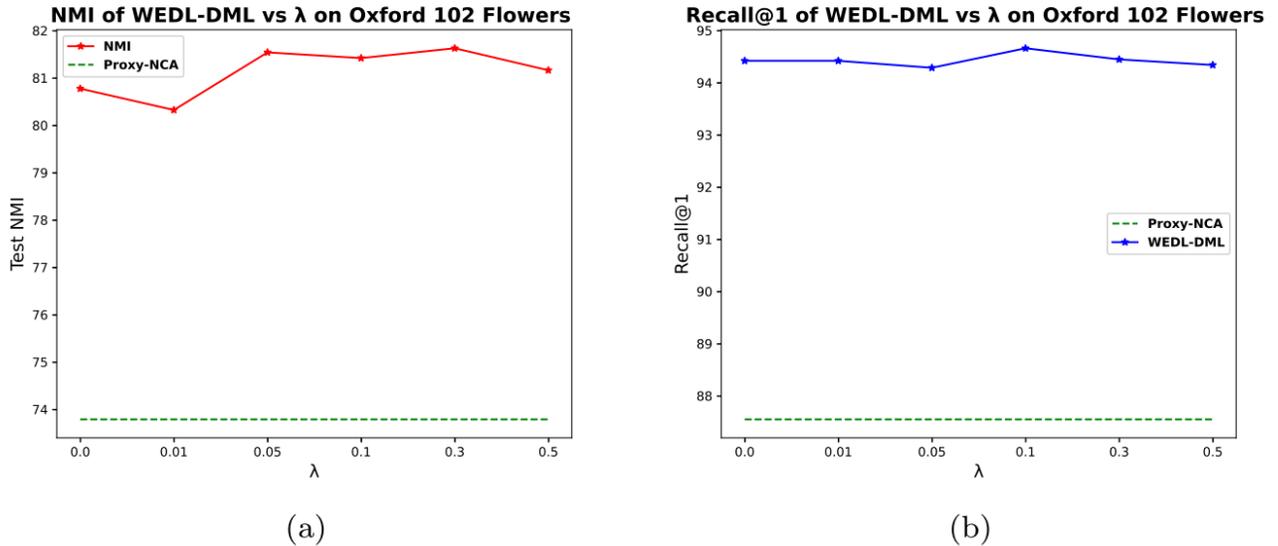

**Figure 12-** Test NMI and Recall@1 of WEDL-DML vs $\lambda$ values on the Oxford 102 Flowers datasets. The results are compared with the Proxy-NCA (best loss) as a baseline.

## 6. Conclusion and Future Work

This work provides a novel approach for combining different loss functions in deep metric learning in a ZSL setting. Here, there is no limitation of selected loss functions, and our methods can work with any set of existing loss functions. For example, in the experiments, we choose loss functions from different categories (i.e., triplet, contrastive, proxy-based, and classification losses).

The loss function values can be at different scales; therefore, we propose a novel rescaling approach based on exponential moving average to normalize the losses. The proposed approach can optimize each loss function as well as its weight in an end-to-end paradigm with no need to adjust any hyper-parameter.

We provide different approaches to combine the loss functions and observe that the best results are obtained when we consider a separate embedding layer for each loss. This method named *WEDL-DML* surpasses the best loss function in all evaluated datasets by a large margin. Besides, we present a novel effective compression distance-based loss function that compresses both the coefficients and embeddings into a single embedding vector. The embedding size of the proposed compressed method (named *WEDL-C*) is identical to each baseline learner. Thus, it is fast as each baseline DML in the evaluation stage. In contrast, its results are very competitive with *WEDL-DML* and indicate large improvement over each baseline loss. The only adjustable hyper-parameter in our method is $\lambda$ (the coefficient of the diversity term) that our results are less sensitive to its selected value.

We perform extensive experiments on some challenging machine vision datasets in a ZSL setting. As the results obtained from different metrics indicate, the proposed methods are consistently outperforming all individual losses. It can be explained by the fact that the selected losses are diverse and focus on different aspects of an optimal semantic embedding. Thus, our effective combining methods yield considerable improvement of the results and generalize well on unseen categories.

In future work, we extend our work for semi-supervised learning. Besides, we aim to evaluate the work on other applications of DML (e.g., person re-identification).

## Acknowledgment

We would like to acknowledge the Machine Learning Lab in the Engineering Faculty of FUM for their kind and technical support.

## Conflict of Interest

The authors have no conflicts of interest to declare that are relevant to the content of this article.

## Availability of data

Datasets used in the experiments are publicly available and can be downloaded from the following links:

1- Oxford 102 Flowers: https://www.robots.ox.ac.uk/~vgg/data/flowers/102/
2- CUB-200-2011: https://github.com/cyizhuo/CUB-200-2011-dataset
3- CARS-196: https://ai.stanford.edu/~jkrause/cars/car_dataset.html